\documentclass{article}
\usepackage{ijcai26}

\usepackage{times}
\usepackage{soul}
\usepackage{url}
\usepackage[hidelinks]{hyperref}
\usepackage[utf8]{inputenc}
\usepackage[small]{caption}
\usepackage{graphicx}
\usepackage{amsmath}
\usepackage{amsthm}
\usepackage{booktabs}
\usepackage{algorithm}
\usepackage{algorithmic}
\usepackage[switch]{lineno}
\usepackage{diagbox}
\usepackage{multirow}
\usepackage{xcolor}
\usepackage{colortbl} 
\usepackage{bbding}

\newcommand{\fzl}[1]{{\color{black}#1}}
\newcommand{\zbj}[1]{{\color{black}#1}}

\title{\emph{EvolveReason}: Self-Evolving Reasoning Paradigm for \fzl{Explainable}  Deepfake \fzl{Facial Image} Identification}

\author{
Binjia Zhou$^{1}$ \footnotemark[2]\and
Dawei Luo$^{3}$ \footnotemark[2]\and
Shuai Chen$^3$\And
Feng Xu$^3$\and
Seow $^3$\and
Haoyuan Li$^2$\and
Jiachi Wang$^1$\and
Jiawen Wang$^3$\and
Zunlei Feng$^{1}$\and
Yijun Bei$^1$ \footnotemark[1]\\
\affiliations
$^1$School of Software Technology, Zhejiang University\\
$^2$College of Computer Science and Technology, Zhejiang University\\
$^3$Ant Group\\
}

\begin{document}

\maketitle
\renewcommand\thefootnote{}
\footnotetext{* Corresponding author.}
\footnotetext[2]{† These authors contributed equally to this work.}
\begin{abstract}
With the rapid advancement of AIGC technology, developing identification methods to address the security challenges posed by deepfakes has become urgent. Face forgery identification techniques can be categorized into two types: traditional classification methods and explainable VLM approaches. The former provides classification results but lacks explanatory ability, while the latter, although capable of providing coarse-grained explanations, often suffers from hallucinations and insufficient detail.
To overcome these limitations, we propose \emph{EvolveReason}, which mimics the reasoning and observational processes of human auditors when identifying face forgeries. By constructing a chain-of-thought dataset, CoT-Face, tailored for advanced VLMs, our approach guides the model to think in a human-like way, prompting it to output reasoning processes and judgment results. This provides practitioners with reliable analysis and helps alleviate hallucination. Additionally, our framework incorporates a forgery latent-space distribution capture module, enabling \emph{EvolveReason} to identify high-frequency forgery cues difficult to extract from the original images.
To further enhance the reliability of textual explanations, we introduce a self-evolution exploration strategy, leveraging reinforcement learning to allow the model to iteratively explore and optimize its textual descriptions in a two-stage process. Experimental results show that \emph{EvolveReason} not only outperforms the current state-of-the-art methods in identification performance but also accurately identifies forgery details and demonstrates generalization capabilities.
\end{abstract}

\section{Introductions}

In recent years, the steady progress of AIGC \cite{shuai2024survey} has made it easy and feasible to generate highly realistic, hard-to-distinguish images. However, the ability to generate high-quality content with just simple prompts and guidance has significantly lowered the threshold for malicious actors to fabricate false information, leading to issues such as bypassing face verification and telephone scams, which pose serious threats to public safety and the authenticity of public opinion. At the same time, it has placed a tremendous burden on human reviewers, making it an urgent task to assist them in effectively identifying sophisticated forgeries.

\begin{figure}[!t]
    \centering
    \includegraphics[width=0.48\textwidth]{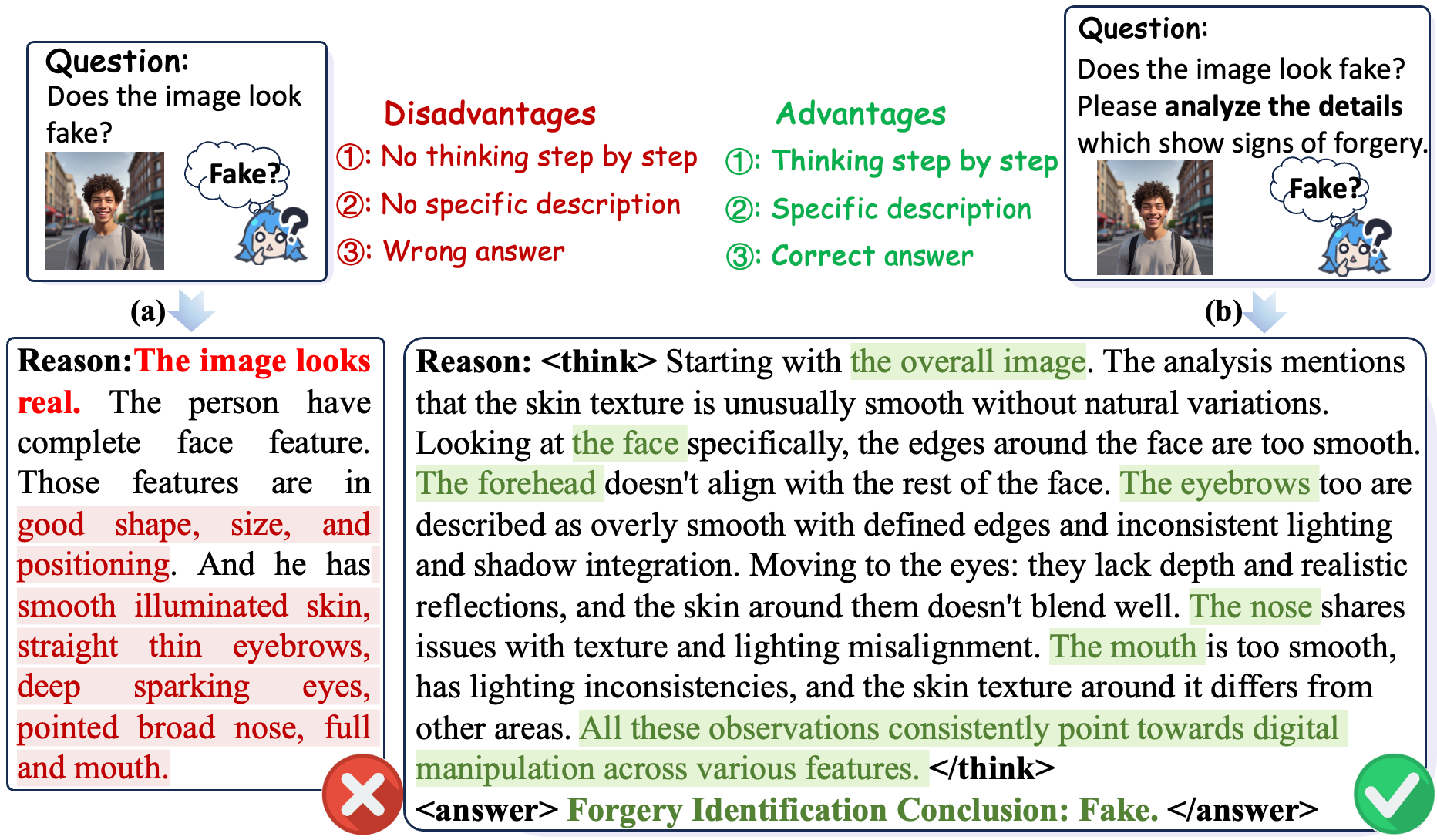}
    \caption{
    \fzl{
    Comparison of our method (b) against other existing explainable approaches (a) in face forgery identification.
    }
    }
    \label{fig:first}
\end{figure}

In order to address the security threats posed by facial image forgery techniques, researchers initially focused on analyzing the facial images themselves to identify inherent visual artifacts in forged facial images~\cite{cao2022end,Dong2022}.
Currently, the main solutions to address the above risks rely on two mainstream approaches, one is based on traditional visual methods, focusing on facial manipulation methods \cite{li2019faceshifter} such as Face2Face \cite{thies2016face2face} and FaceSwap \cite{korshunova2017fast}. However, relying solely on a visual modality for forgery detection treats the process as a black box, yielding only binary classification without critical visual cues and understanding the basis of the discrimination. Although segmentation-based methods can locate tampered regions via masks to provide visual explanations~\cite{sun2025towards}, they still can't generate natural language explanations, making them not friendly enough in understanding for reviewers.

To enhance the explanatory ability and informational depth of facial image forgery identification models, some researchers have endeavored to incorporate textual information, developing multi-modal deepfake identification frameworks. Leveraging established text-image encoding technologies such as CLIP~\cite{clip} and BLIP~\cite{blip}, a bridge for feature interaction between the image and text modalities has been constructed. Another attempt to overcome this uses vision and language models (VLM) \cite{zhang2025common,kundu2025truthlens,song2024learning}. New approaches have experimented with pre-trained VLM, for example GPT 4o \cite{hurst2024gpt}, to detect forged content while offering explanations that are easy for humans to follow, addressing challenges of accuracy and explanatory ability in deepfake identification. Although certain explanations can be provided, hallucinations remain a significant issue, counteracting the intended effect. Moreover, the datasets used in VLM-based methods, such as DD-VQA~\cite{zhang2025common}, contain a large amount of noise, posing considerable challenges for training VLM.

To address these challenges, we introduce \emph{EvolveReason}, a novel self-evolving reasoning multimodal explainable framework. For forged explanatory ability identification, \emph{EvolveReason} is trained on a newly chain-of-thoughts dataset, CoT-Face, and in order to ensure the accuracy of forgery identification and textual explanations, we have designed three module:
a) To capture high-frequency forgery details that are difficult to identify in the RGB image, the forgery visual clue extraction enriches the downstream visual inputs by providing reconstructed differences and fourier frequency domain data. 
b) Based on the visual forgery clues, initial cot alignment integrates the anti-forgery observation reasoning into the VLM using the constructed CoT-Face dataset, allowing the model to initially develop the ability to explain forgery clues.
c) To ensure the reliability of explanations of the text, self-evolving reasoning drives the VLM to explore answers that surpass human-provided labels through reinforcement learning, directly addressing and explaining the forgery traces of different regions. 
Through extensive experiments, we have demonstrated that our method achieves high forgery identification accuracy and provides reliable textual explanations for the forgery clues. The contributions of this work are as follows:

\begin{itemize}
    \item We propose the \emph{EvolveReason} framework, which addresses the issue of noise caused by mismatches between visual forgeries and textual descriptions, and enables the VLM to emulate human reviewers in observing forged images from a global perspective to key local details, achieving explainable deepfake identification.

    \item The proposed self-evolving reasoning strategy employs a reinforcement learning-based reward mechanism to drive the VLM to explore the best thought and output, combined with the proposed distribution consistency constraint, effectively improve the model's forgery identification performance and text reliability.
    
    \item We have constructed a new chain-of-thoughts dataset called CoT-Face, which contains over 5,900 samples. Each sample includes multiple forgery traces, from overall to local details, in forged images, laying the groundwork for training the model to adopt the forgery identification approach of human reviewers.
    
    
\end{itemize}

\section{Related Work}
\subsection{Conventional Face Forgery  Methods}
Mainstream approaches in this field rely on visual signal capture, traditional techniques magnify forgery traces in the frequency~\cite{qian2020thinking,luo2021generalizing} and spatial domains~\cite{bei2024large,li2020face}, enabling precise identification of subtle manipulations and facial artifacts. However, the rise of AIGC-style global forgeries has exposed the limitations of detectors that focus on local image regions. To meet this new challenge, a wave of GAN-based~\cite{huang2022fakelocator,guarnera2020deepfake} and diffusion-based~\cite{somepalli2023diffusion} methods has emerged, achieving state-of-the-art performance in fine-detail analysis and cross-dataset generalization: for example, CNNDetection~\cite{wang2020cnn} trained on ProGAN~\cite{karras2017progressive} exhibits remarkable generalization, while DIRE~\cite{wang2023dire} effectively addresses diffusion-generated fakes because genuine images themselves cannot be perfectly reconstructed. Nevertheless, in complex open-world settings these methods still show restricted generalization. Moreover, they usually frame the task as binary classification, yielding only two-class outputs or coarse heatmaps that provide little insight into the model’s decision process~\cite{chen2024diffusionfake,kundu2025towards}. \citeauthor{sun2025towards} attempts to enhance accuracy and visual explanation through masking, but downstream explanations using general large models cannot succinctly pinpoint forgery traces in text form. Therefore, it is essential to develop an open-world face-forgery analysis model that is both highly generalizable and robust, while also delivering user-friendly and explainable results under realistic conditions across diverse real datasets.

\begin{figure*}[!t]
  \centering
  \includegraphics[width=0.98\textwidth]{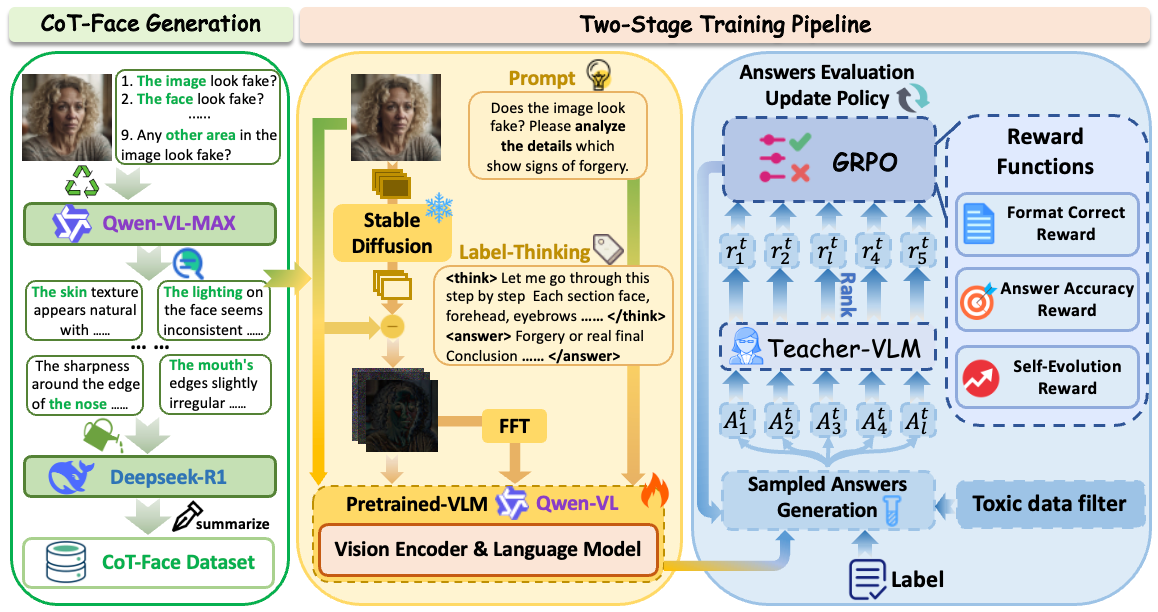}
  \caption{
    The framework of the proposed \emph{EvolveReason}. During training, the forgery distribution latent space feature extraction is used to enhance high-frequency forgery clues in the input of forged images and restore detailed information. Subsequently, fine-tuning the VLM with the forgery identification knowledge guided by the chain-of-thought in the CoT-Face dataset encourages the model to mimic the real auditor's process of observing and identifying forged images. Finally, reinforcement learning is employed to self-evolve the VLM's text explanation ability, enhancing the reliability of explanations and identification.
  }
  \label{fig:main}
\end{figure*}

\subsection{VLM-based Explainable}
Text-based deepfake explainability offers finer-grained insights than traditional visualization techniques~\cite{huang2024ffaa,chang2023antifakeprompt,gao2025fakereasoning}, because it employs natural language to spell out specific inconsistencies or manipulations in a way people can readily follow. Early attempts simply used general-purpose VLM~\cite{jia2024can}, zero-shot identification with ChatGPT~\cite{achiam2023gpt} but these models lack forgery-specific knowledge from pre-training, so their accuracy and generalization were limited.
DD-VQA~\cite{zhang2025common} introduced a VQA-style dataset built on FaceForensics++~\cite{rossler2019faceforensics++}, laying the groundwork for training dedicated, explainable large models in this domain. Based on that dataset, CorrDetail~\cite{zhou2025corr} reinforced the sensitivity of LLaVA’s image-features and cross-domain generalization through a self-correcting Q\&A scheme. TruthLens~\cite{kundu2025truthlens} went further, injecting the fine-grained local feature perception of vision encoder such as DINOv2~\cite{oquab2023dinov2} into VLM to heighten awareness of subtle forgeries. \citeauthor{sun2025towards} tackled inherent hallucination issues during explanation, it first masks forged regions for coarse localization and then fine-tunes a VLM with a comprehensive prompting strategy and structured annotations.
By contrast, \emph{EvolveReason} not only introduces a brand-new CoT-Face dataset tailored to face forgery analysis but also fine-tunes a VLM guided by correct reasoning chains. The result is precise, fine-grained forgery identification that delivers step-by-step explanations, from global assessment down to local cues, making the model’s decision logic transparent to end users.

\subsection{Explainable Model with Chain-of-Thoughts}
Similar to Chain-of-Thought (CoT) prompting in conventional large language models~\cite{wang2024t,mitra2024compositional}, reasoning prompts for VLM fall into two broad categories. The first enriches the model with additional image-related cues to strengthen object recognition, as in Set-of-Marks~\cite{yang2023set} and ICot~\cite{gao2025interleaved}.
The second guides reasoning through textual logic: DDCoT~\cite{zheng2023ddcot} breaks a hard problem into a sequence of simpler sub-problems that the model solves one by one before synthesizing the final answer, while CCoT~\cite{mitra2024compositional} asks the VLM to describe every object in the image to support subsequent reasoning.
Yet none of these approaches provides a concrete tutorial that teaches the model how to observe an image accurately for a given downstream task.
To fill this gap, we introduce a domain-specific CoT framework for face-forgery identification. Our method incrementally steers the VLM through the verification process and uses reinforcement learning to effectively prune redundant or repetitive content, thereby yielding more concise, readable, and explainable outputs across diverse scenarios.

\section{CoT-Face Dataset}
To address the challenges arising from the lack of proper guidance methods and corresponding datasets for effectively steering the reasoning process of VLMs in Face Forgery identification, Fig.\ref{fig:main} demonstrates how we constructed a specialized dataset named CoT-Face tailored for the current domain, using high-parameter multi-modal large models and the FF++ dataset~\cite{rossler2019faceforensics++}. Specifically, we employed Qwen-72B-VL-MAX~\cite{wang2024qwen2} to iteratively address multiple detailed queries, decomposing the complex overall judgment task into separate, key-region-specific questions. Subsequently, leveraging Deepseek-R1's~\cite{guo2025deepseek} strong reasoning and summarization capabilities, we integrated the iterative results from the previous phase and fed them into a downstream LLM. Guided by specific instructions, this model condensed and refined the results into an initial dataset comprising approximately 6,000 samples. Finally, professional forgery reviewers conducted a second round of verification, assessing everything from the smoothness of local details to the integrity of the overall structure, the correctness of the conclusions, and the precision of the descriptions, thereby refining and denoising the data. This process ultimately yielded a compact, high-quality CoT dataset suitable for training smaller VLM models, laying a solid foundation for subsequent downstream tasks. 

\zbj{We provide the reviewers’ expertise and evaluation logic, along with detailed explanations of the methodology and dataset structure, in the \emph{supplementary material}}.

\section{Method}
Fig.\ref{fig:main} illustrates the overall framework of \emph{EvolveReason}.
The first subsection presents the forgery visual clue extraction module, which uses diffusion models to provide richer visual cues for detecting fine-grained facial forgeries, enhancing the VLM’s visual-feature extraction.
The second subsection explains how the CoT-Face dataset drives the VLM’s forgery-detection reasoning and steers its chain-of-thought process.
The final subsection applies group relative policy optimization (GRPO) to fine-tune the VLM, optimizing the reward function to allow the model to autonomously evolve and explore textual descriptions beyond human annotations.

\subsection{Forgery Visual Clue Extraction (FVCE)}

Most prior VLM-based approaches deliver additional image clues to the model through text prompts, yet this modality is not sufficiently congenial: forgery details are often subtle, so purely textual descriptions struggle to meet identification requirements. We therefore enrich the model’s visual input to strengthen its feature-capture capability.

Concretely, we first feed the original image
$I$ into a pre-trained Stable Diffusion and extract restored images $R_n$,
\begin{equation*}
    {R_{1},R_{2},...,R_n,...,R_{N}}=Diffusion(I), 
\end{equation*}
from $N$ time steps, that show the most pronounced reconstruction effects. The last $K$ restored image is then subtracted from the original to produce a set of difference images $D_n$ \fzl{as follows:}
\begin{equation*}
    D_n = I - R_n, \ \ n\in (N,N-1,\ldots,N-K).
\end{equation*}
This series preserves the coarse-to-fine restoration cues, exposing abundant structural information and local details. Because image forgeries are largely reflected in high-frequency components, for example, techniques such as Face2Face and FaceSwap produce abrupt pixel-level changes, capturing this property allows the model to detect subtle features that would otherwise be hard to observe. Finally, we apply a Fourier transform to every $D_n$ to obtain the corresponding frequency domain images $F_n$ \fzl{as follows:}
\begin{equation*}
    F_n = FFT(D_n),\ \ n\in (N,N-1,\ldots,N-K).
\end{equation*}
With these pre-processing steps completed, all image data are ready for downstream processing. 

\subsection{Initial Cot Alignment (ICA)}
Most current VLM-based forgery identification methods give only a binary verdict or a very rough explanation. This training paradigm turns the model into a black box: we cannot see whether it has truly learned a sound reasoning process, and human reviewers struggle to assess the reliability of its outputs. To overcome these issues, we apply special measures during the fine-tuning and alignment stage.
First, at the system-prompt level we instruct the VLM to enclose its reasoning in \textless think\textgreater …\textless /think\textgreater  tags and its final answer in \textless answer\textgreater … \textless /answer \textgreater tags, this is the basic output format. The content inside these tags represents the forgery identification knowledge the model must learn during fine-tuning and alignment. In addition, we supply a structured answer template, for example, we require the model to list detected forgery clues in a logical sequence for each region (Overall image, Face, Eyebrows, etc.). We reinforce these capabilities by concatenating extra visual information and modalities along the channel dimension to better suit the forgery identification task. The specific formula is as follows:
\begin{equation*}
    ExtraInfo = Concat(\sum\nolimits_{n=N-K}^{N}F_n,\sum\nolimits_{n=N-K}^{N}D_n),
\end{equation*}
\begin{equation*}
    Answer = VLM(I,ExtraInfo,Prompt,Label),
\end{equation*}
where $I$, $F_n$, and $D_n$ are all generated by the FVCE module, the $Prompt$ is the standard query, such as “Does the image look fake?” and the $Label$ is taken directly from the dataset’s own annotated input.

\begin{figure}[!t]
    \centering
    \includegraphics[width=0.5\textwidth]{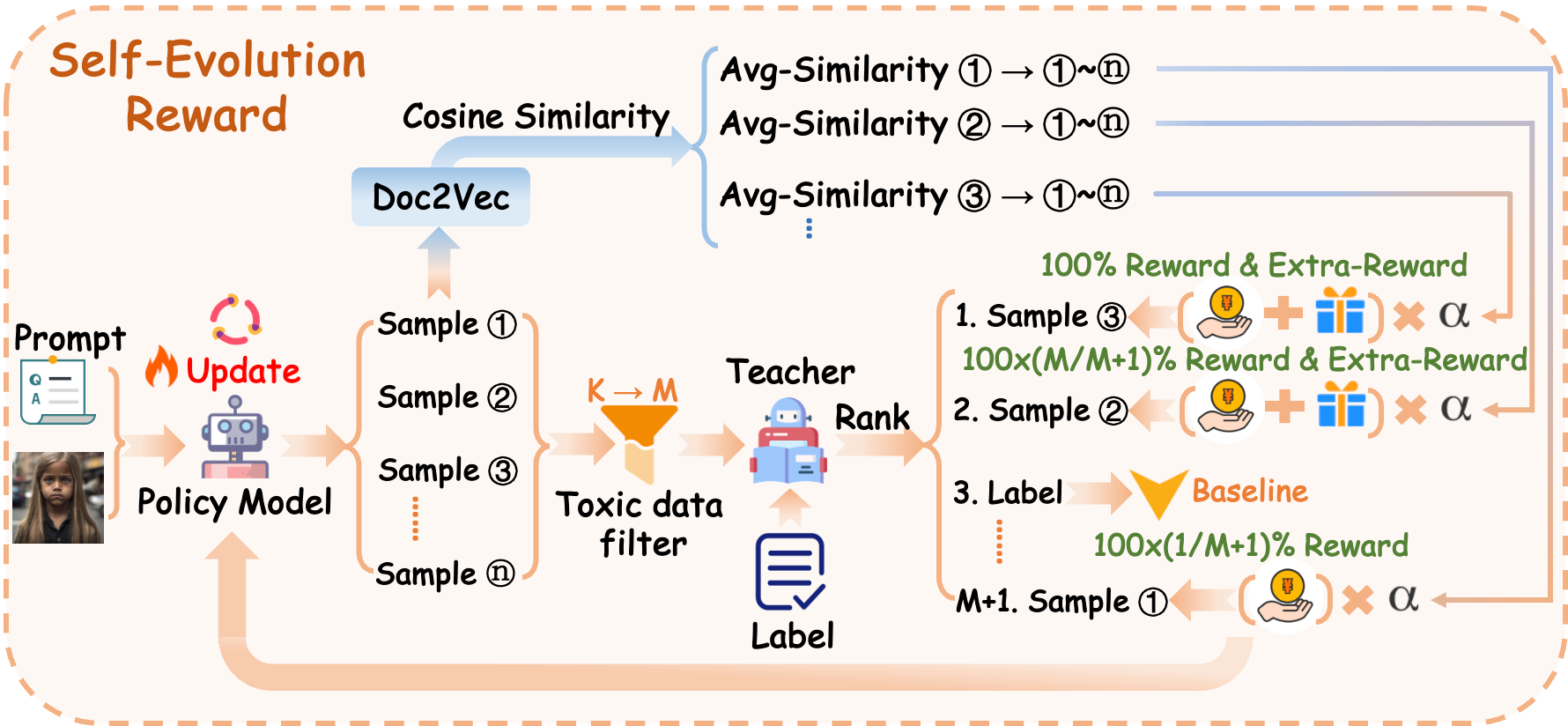}
    \caption{
    \zbj{The overall workflow and schematic diagram of self-evolution reasoning.}
    }
    \label{fig:grpo}
    \vspace{-0.5em}
\end{figure}

\begin{figure*}[!t]
  \centering
  \includegraphics[width=0.98\textwidth]{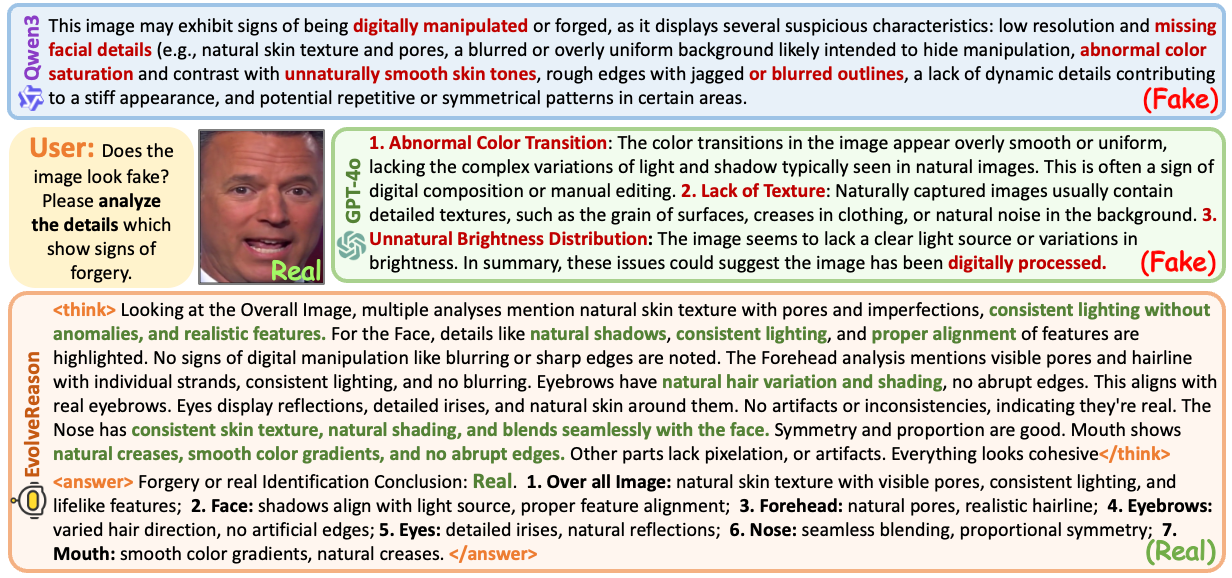}
  \caption{
  Qualitative examples of results on certain samples. For
  failure cases of Qwen3-235B-A22B and GPT-4o, \emph{EvolveReason} achieves correct classification results along with corresponding explanations.
  }
  \label{fig:vis}
\end{figure*}

\subsection{Self-Evolving Reasoning (SER)}

There has already been extensive application of reinforcement learning methods, such as GRPO, in general VLMs, which have \fzl{verified} effectiveness in helping models explore correct reasoning paths and improve performance through self-feedback across diverse settings. However, attempts to leverage these methods in face-forgery identification remain limited. Therefore, we propose a novel textual-exploration reward function that not only encourages the model to seek “aha” moments~\cite{guo2025deepseek} beyond the constraints of existing labels, but also leverages the answer distribution from multiple GRPO-based samplings to prudently constrain the model’s progression, as detailed in Fig.~\ref{fig:grpo}.

For a given prompt $Q$, VLM samples $C$ candidate responses, we use the first version of \emph{EvolveReason}, which has already acquired forgery identification knowledge, as a toxic data filter to eliminate other low-quality generated data, retaining only the top $M$ responses, denoted as $\{A_m\}^M_{m=1}$, sampled from the policy $\pi_{\theta}$. We then evaluate each response using a reward function $R_m(Q,A_m)$, which measures the quality of the candidate response in the context of the given question. To determine the relative merits of these responses, GRPO normalizes the rewards by computing their mean and standard deviation, and then derives the advantage $Adv$ as follows: 
\begin{equation*}
    Adv=\frac{R-\frac{1}{M}\sum_{m=1}^{M}R_m}{\sqrt{\frac{1}{M-1}\sum_{m=1}^{M}\left(R_m - \frac{1}{M}\sum_{p=1}^{M}R_p\right)^2}}.
\end{equation*}
The goal of our GRPO is to explore and identify the sampled response with the highest advantage value, while the reward function is designed to align with human preferences.
Accordingly, across $T$ iterations we continually allow the VLM’s outputs to undergo self-evolution, aiming to produce the text that most accurately captures the forgery cues, which can be expressed as follows:
\begin{align*}
\bigl(A_1^{t},\dots,A_M^{t}\bigr)
  = GRPO\bigl(A^{t-1}\bigr),
\end{align*}
\begin{equation*}
    A^{t}=TeacherVLM(A_{1}^{t},A_{2}^{t},\dots,A_{M}^{t},\hat{A}^t),
\end{equation*}
where $\hat{A}^t$ means the ground-truth in specific $t$ iteration and $t \in (2,3,\dots,T)$, and we employ Qwen-72B-VL-MAX as the TeacherVLM. In this paper, we adopt three rule-based approach for the reward function:
\begin{equation*}
    R_{all}=R_{fmt}+R_{acc}+R_{see}.
\end{equation*}

\textbf{Format Reward:} The output answers are evaluated with two parts:
\begin{equation*}
    R_{fmt}=R_{tag}+R_{key}.
\end{equation*}
If the response contains specific tags, a reward is given \fzl{as follows}:
\begin{equation*}
    R_{tag} =0.5\ \ if \ \ match(A_i, tags) \ \ else \ \ 0.0,
\end{equation*}
\begin{equation*}
    tags = \text{``\textless} think\text{\textgreater}..\text{\textless} /think\text{\textgreater \textless} answer\text{\textgreater ..\textless}/answer\text{\textgreater"}.
\end{equation*}
If the response further contains keywords that match the keywords$\in \{Overall \ image,Face,Nose,...,Neck\}$, additional rewards are incrementally granted:
\begin{equation*}
    R_{key}=1.0\ \ if\ \ keywords \in A \ \ else \ \ 0.0  .
\end{equation*}

\textbf{Accuracy Reward: }This portion of the reward is determined mainly by whether the model’s output matches the label in a binary classification sense, if they coincide, the reward is granted:
\begin{equation*}
    R_{acc}= 1.0\ \ if \ \ Cls(A)==Cls(\hat{A})\ \ else \ \ 0.0 ,
\end{equation*}
where $Cls(\cdot)$ denote their respective classification outcomes for the potentially forged image, either real or forgery.

\textbf{Self-Evolution Reward:} This reward term introduces a  Teacher VLM that ranks the generated samples by their image–text alignment over multiple samplings $Rank(A_1,A_2,...,A_M,\hat{A})$. Uniquely, before the ranking step we treat the ground-truth label $\hat{A}$ as an additional sample and include it in the candidate pool, 
after the ranking step, we assign rewards according to each sample’s position. Concretely, with the ground-truth label as our reference point, any sample that ranks above the label receives an extra bonus, encouraging the model to boost the probability of these “beyond-upper-bound” explorations instead of merely gravitating toward the label. Samples that rank below the label receive only a baseline reward. Formally, this can be written as follows:
\begin{equation*}
    R_{see}=
    \begin{cases}
        (\beta*\frac{M-r}{M}+e^{1-\frac{A \cdot \hat{A}}{||A||_2 \ ||\hat{A}||_2}})*\alpha, \ r <\hat{r} \\
        (\beta*\frac{M-r}{M})*\alpha, \ r>\hat{r}
    \end{cases},
\end{equation*}
where $r$ and $\hat{r}$ represents the rank of specific sample $A$ and $\hat{A}$, the $\beta$ is a hyperparameter, that is the reward weight we set to $1.5$. Specifically, because we do not want these breakthrough cases to push the model’s exploration range too far, we monitor the distribution of multiple samples and check whether the outputs converge. 
If the distribution is still widely dispersed, it suggests the model has not genuinely captured the image-forgery cues and is instead hallucinating. 
\fzl{Therefore, we introduce a coefficient $\alpha$: }
\begin{equation*}
    \alpha= \frac{1}{M}\sum\nolimits_{m=1}^{M}\frac{A \cdot A_m}{||A||_2 \ ||A_m||_2}, 
\end{equation*}
\fzl{
which controls the model's learning step size for capturing image-forgery cues while preventing hallucination.
}

\begin{table*}[t]
\centering
\renewcommand{\arraystretch}{1.2}
\setlength{\tabcolsep}{11pt}
\resizebox{\textwidth}{!}{
\begin{tabular}{c|ccccc|ccccc|ccc}
\hline
\multirow{3}{*}{\diagbox[width=14em, font=\Large\centering]{Methods}{Datasets}}
& \multicolumn{10}{c|}{Intra-Testing}
& \multicolumn{3}{c}{Cross-Testing}
\\ \cline{2-14}
& \multicolumn{5}{c|}{FF++ (LQ)}
& \multicolumn{5}{c|}{FF++ (HQ)}
& \multicolumn{3}{c}{CelebDF (trained on FF++ (LQ))}
\\ \cline{2-14}
& ACC↑ & AUC↑ & CIDEr↑ & SPICE↑ & Top-2↑
& ACC↑ & AUC↑ & CIDEr↑ & SPICE↑ & Top-2↑
& ACC↑ & AUC↑ & EER↓
\\ \hline
XceptionNet~\cite{rossler2019faceforensics++}
& 86.86 & 93.50 & / & / & /
& 95.04 & 96.30 & / & / & /
& 60.11 & 61.80 & 41.73
\\
RFM~\cite{wang2021representative}
& 87.06 & 89.83 & / & / & /
& 95.69 & 98.79 & / & / & /
& 64.42 & 65.64 & 38.80
\\
RECCE~\cite{cao2022end}
& 91.03 & 95.02 & / & / & /
& 97.06 & 99.32 & / & / & /
& 67.96 & 68.71 & 35.73
\\
UIA-VIT~\cite{zhuang2022uia}
& 91.05 & 94.88 & / & / & /
& 98.62 & 99.33 & / & / & /
& 69.80 & 70.15 & 35.82
\\
HiFi-Net~\cite{guo2023hierarchical}
& 89.25 & 92.10 & / & / & /
& 97.41 & 98.56 & / & / & /
& 67.20 & 68.80 & 36.13
\\
PFG-DD~\cite{lin2024preserving}
& 91.03 & 93.47 & / & / & /
& 97.33 & 98.68 & / & / & /
& 68.51 & 69.68 & 35.94
\\
Forensics Adapter~\cite{cui2025forensics}
& 93.56 & 96.20 & / & / & /
& 98.76 & 99.44 & / & / & /
& 73.37 & 74.55 & 33.06
\\
CorrDetail~\cite{zhou2025corr}
& \underline{94.41} & \underline{96.93} & / & / & /
& \underline{99.28} & \textbf{99.92} & / & / & /
& 72.32 & 72.80 & 33.87
\\
RECCE (DD-VQA)~\cite{zhang2025common}
& 92.08 & 95.36 & 2.06 & \underline{0.69} & \underline{34.6}
& 98.65 & 99.79 & \underline{2.43} & \underline{0.77} & \underline{42.2}
& 69.46 & 70.21 & 35.63
\\
FFAA~\cite{huang2024ffaa}
& 92.48 & 94.74 & 1.92 & 0.62 & 21.4
& 98.68 & 99.02 & 2.28 & 0.70 & 16.8
& \underline{74.08} & 75.31 & 30.95
\\
FakeReasoning~\cite{gao2025fakereasoning}
& 92.79 & 95.51 & \underline{2.14} & 0.66 & 21.5
& 98.71 & 99.78 & 2.25 & 0.74 & 25.6
& 72.68 & 73.94 & 31.07
\\
SIDA~\cite{huang2025sida}
& 92.13 & 94.32 & 1.83 & 0.59 & 16.6
& 97.82 & 98.92 & 2.19 & 0.68 & 19.2
& 73.00 & 74.39 & \underline{30.58}
\\
\cite{sun2025towards}
& 91.86 & 94.07 & 1.97 & 0.60 & 20.4
& 98.97 & 99.16 & 2.31 & 0.74 & 27.4
& 73.91 & \underline{76.87} & 31.38
\\ \hline \rowcolor{lightgray!40}
\textbf{\emph{EvolveReason} (ours)}
& \textbf{95.01} & \textbf{97.04} & \textbf{2.29} & \textbf{0.75} & \textbf{81.4}
& \textbf{99.40} & \underline{99.88} & \textbf{2.51} & \textbf{0.81} & \textbf{73.4}
& \textbf{76.50} & \textbf{78.41} & \textbf{28.40}
\\ \hline
\end{tabular}
}
\caption{
Comparison of \emph{EvolveReason} with SOTA methods across both intra- and cross-datasets. 
The highest and second-highest performances are highlighted in \textbf{best} and \underline{second best}, respectively. And / denotes methods that fail to generate interpretable text. Moreover,
ACC and AUC denote that higher values are preferred, while EER  signifies that lower values are desired.
All scores are expressed in \%.
}
\label{tab:comparison}
\end{table*}

\section{Experiments}
In this section, we introduce the experimental setup, present a performance comparison between \emph{EvolveReason} and recent approaches across various datasets. 

\textbf{Architecture and Implementation Details.} 
During the CoT-Face dataset construction phase, we queried Qwen-72B-VL-MAX to generate rationales and employed Deepseek-R1 to condense and refine them. For fine-tuning, we extracted distributions with Stable Diffusion v1.5 and used Qwen2.5-VL-7B as the base model. Training on the CoT-Face dataset was carried out with a learning rate of 2$e^{-5}$, for 4 epochs, using a batch size of 16 on NVIDIA A100 GPUs. 

More details (e.g., FVCE sensitivity analysis, SER module’s sampling strategy, time-efficiency analysis, and hyperparameter settings) are provided in \emph{ supplementary material.}

\textbf{CoT-Face Dataset.} Based on the original data from the FF++ dataset, we have constructed the CoT-Face dataset,
which consists of 5953 [image, question, answer] triples to
support the chain-of-thoughts fine-tuning and SER. The dataset includes not only the condensed outputs and their corresponding prompts, but the results from every generation stage polling step, making it easy to extend the corpus further, each example contains ten such polling iterations.

\begin{table}[!t]
\centering
\renewcommand{\arraystretch}{0.95} 
\setlength{\tabcolsep}{4.5pt}      
\resizebox{0.4\textwidth}{!}{
\small 
\begin{tabular}{@{}ccc|cc|cc@{}}
\hline
\multicolumn{3}{c|}{Module} & \multicolumn{2}{c|}{Intra-Testing} & \multicolumn{2}{c}{Cross-Testing} \\
\hline
\multirow{1}{*}{FVCE} & \multirow{1}{*}{ICA} & \multirow{1}{*}{SER} & \multicolumn{1}{c}{LQ} & \multicolumn{1}{c|}{HQ} & \multicolumn{1}{c|}{CelebDF} & \multicolumn{1}{c}{DFDC} \\
\hline
\XSolidBrush   & \XSolidBrush   & \XSolidBrush    & 88.44 & 91.87 & 65.58 & 69.42 \\
\XSolidBrush   & \XSolidBrush   & \CheckmarkBold  & 91.91 & 94.59 & 71.03 & 74.97 \\
\XSolidBrush   & \CheckmarkBold & \XSolidBrush    & 92.75 & 96.15 & 73.11 & 76.59 \\
\CheckmarkBold & \XSolidBrush   & \XSolidBrush    & 91.20 & 94.81 & 68.59 & 74.31 \\
\XSolidBrush   & \CheckmarkBold & \CheckmarkBold  & 96.11 & 98.43 & 77.39 & 78.28 \\
\CheckmarkBold & \XSolidBrush   & \CheckmarkBold  & 93.80 & 96.72 & 74.07 & 76.80 \\
\CheckmarkBold & \CheckmarkBold & \XSolidBrush    & 95.92 & 97.38 & 75.20 & 78.51 \\
\rowcolor{lightgray!40}
\CheckmarkBold & \CheckmarkBold & \CheckmarkBold  & \textbf{97.04} & \textbf{99.88} & \textbf{78.41} & \textbf{80.94} \\
\hline
\end{tabular}}
\caption{Ablation study of \emph{EvolveReason} with FF++ datasets. Cross-domain testing is conducted on LQ. Best are highlighted in \textbf{bold}. All numbers are AUC (\%).}
\label{tab:abl}
\vspace{-1.5em}
\end{table}

\textbf{Evaluation Dataset and Metrics.} We evaluate the method using the FF++~\cite{rossler2019faceforensics++}, Celebf~\cite{li2020celeb}, DFD, DFDC~\cite{dolhansky2020deepfake} and DeepFaceGen~\cite{bei2024large} datasets. The DFD corpus comprises more than 363 original clips, and beyond the footage it includes over 3,000 processed videos generated with the deepFake technique. The DFDC dataset comprises more than $100,000$ videos produced with eight forgery techniques. CelebDF relies exclusively on face swapping and contains 5,639 manipulated clips. FF++ contributes 4,000 videos generated with four techniques Deepfake, Face2Face, FaceSwap, and NeuralTextures. For both datasets, each video is cropped to face images, creating two of the field’s most common image-level benchmarks. DeepFaceGen offers both video and still-image forgeries, from which we collect forged faces produced by 27 methods, spanning conventional task-driven pipelines and our new prompt guided approach. Following prior benchmark practice, we report Accuracy (ACC), Equal Error Rate (EER), and Area Under the Curve (AUC).

\zbj{More experimental on DFD and DFDC, as well as analyses of failure cases, are provided in the \emph{ supplementary material.}}

\subsection{Comparison with SOTAs} 
We compare \emph{EvolveReason} with mainstream approaches. These include traditional binary classifiers—XceptionNet, EN-B4, RFM, RECCE, UIA-ViT, HiFi-Net, PFG-DD, and Forensics Adapter—as well as VLM-based explainability methods such as CorrDetail, RECCE (DD-VQA), FFAA, FakeReasoning, and SIDA.
We fine-tuned \emph{EvolveReason} on the CoT-Face dataset constructed from FF++. As reported in Table.\ref{tab:comparison}, our model delivers the best overall performance.

\zbj{We compare the quality of text generation using the SPICE~\cite{spice2016} and CIDEr~\cite{vedantam2015cider} metrics.} Moreover, we employed ChatGPT-o3 to rank the explainable methods’ outputs on samples according to their image–text consistency. For each of 500 FF++ samples, we selected the top two methods; \emph{EvolveReason} emerged most frequently in rankings. Fig.\ref{fig:vis} provides a visual comparison of the explanations produced by \emph{EvolveReason} and a generic VLM in the context of forgery identification.

\begin{figure*}[!t]
  \centering
  \includegraphics[width=0.87\textwidth]{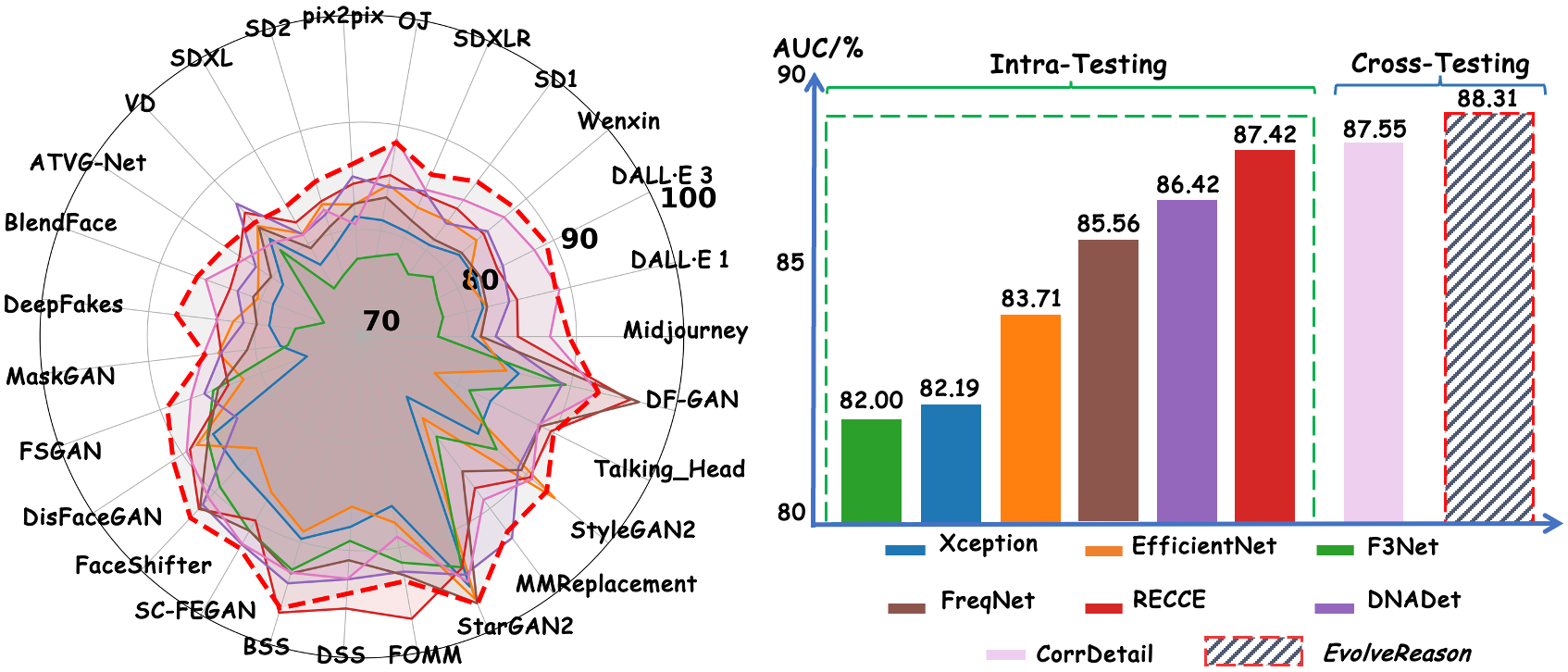}
  \caption{
  \zbj{Generalization performance of \emph{EvolveReason} versus the in-domain performance of existing methods. The radar chart illustrates identification performance across different forgery categories: solid lines represent baseline methods, while the dashed line corresponds to \emph{EvolveReason}. The bar chart reports mean AUC for each approach. All six baseline methods were trained and evaluated on DeepFaceGen, whereas CorrDetail and \emph{EvolveReason} were trained on FF++ and tested on DeepFaceGen, representing a cross-dataset evaluation.}
  }
  \label{fig:faceGen}
\end{figure*}

\subsection{Generalization on Different Datasets.} 
To further validate the effectiveness of \emph{EvolveReason}, we designed experiments that probe its ability to generalize. On the CelebDF dataset, we first trained \emph{EvolveReason} on CoT-Face and then transferred it to CelebDF for cross-dataset evaluation, comparing its performance with other methods also trained on FF++.
For DeepFaceGen, we contrasted \emph{EvolveReason} with seven state-of-the-art identification models. Notably, six of these models were tested in an intra-dataset manner—trained and evaluated on DeepFaceGen, whereas CorrDetail and \emph{EvolveReason} were trained on FF++ and then tested on DeepFaceGen, constituting a cross-dataset setting.
The results on CelebDF are summarized in Table.\ref{tab:comparison}. In the cross-dataset experiments, \emph{EvolveReason} delivered the best performance, surpassing all competing approaches. On the more challenging DeepFaceGen benchmark (Fig.\ref{fig:faceGen}), even under the less favorable scenario in which \emph{EvolveReason} is trained only on CoT-Face but tested on DeepFaceGen, while all comparison methods undergo same-dataset training and testing—\emph{EvolveReason} still outperforms the SOTA model.

\subsection{Ablation Study} 
\textbf{Illustrating the Effectiveness of FVCE.} Table.\ref{tab:abl} demonstrates that the module, capturing high-frequency forgery cues and extracting latent-space features, boosts performance on both FF++ and CelebDF, in a consistent and significant manner, showing strong cross-domain generalization. This indicates that FVCE helps the base model tackle difficult, hard-to-spot samples at the raw-input level. We present the visualization results of the step-by-step restoration difference in the \emph{supplementary material} to further demonstrate the effectiveness of the diffusion mechanism in capturing high-frequency forgery cues.

\textbf{Discussing the influence of ICA.} To assess the performance gains brought by the chain-of-thought strategy, we compared \emph{EvolveReason} with a generic VLM in terms of forgery identification capability. Besides measuring binary classification accuracy, we also examined whether the responses are sufficiently fine-grained and whether they provide reviewers with precise explanations of forged clues. As illustrated in Fig.\ref{fig:vis}, our method accurately detects forged images in several cases where the generic VLM fails, while also outputting its reasoning process from global overview down to local details step-by-step describing potential forgery artifacts in each key region, and concluding with a concise, well-focused summary in answer section.

\textbf{Observe the practicality of SER.} To examine how the SER module enables self-evolution in \emph{EvolveReason}’s textual outputs, we visualized the model’s responses on a CoT-Face sample before and after reinforcement. Prior to reinforcement, the model underwent fine-tuning with knowledge injection; it produced rigid, template-like answers dictated by the prompt. It still reported fabricated details about the “Neck” area even when the forged image contained little to no neck region, and omitted other relevant cues, which is sub-optimal. After reinforcement, however, the model first evaluates whether each key region merits a response and provides more granular descriptions for those exhibiting forgery artifacts. Its coverage of facial regions is more comprehensive and detailed, aligning better with auditors’ needs.
Ablation visualizations are provided in the \emph{supplementary materials}.

\section{Conclusion}
In this paper, we propose \emph{EvolveReason}, a new face-forgery identification framework that boosts both performance and explanatory ability through chain-of-thought training and guided forgery reasoning. The FVCE module captures subtle high-frequency artifacts and cues hard to spot in manipulated images, while ICA injects reasoning-driven forgery knowledge into the VLM, producing explanations auditors can readily understand. Finally, SER introduces a reinforcement-learning-based self-evolution strategy that overcomes the ceiling imposed by manual labels and expands the model’s capacity for textual exploration. Across multiple benchmarks, \emph{EvolveReason} surpasses recent state-of-the-art methods with higher accuracy and generalization.

\bibliographystyle{named}
\bibliography{ijcai26}

\end{document}